\documentclass[letterpaper]{article} 
\usepackage{aaai2026}
\usepackage{times}  
\usepackage{helvet}  
\usepackage{courier}  
\usepackage[hyphens]{url}  
\usepackage{graphicx} 
\usepackage{natbib}  
\usepackage{caption} 
\usepackage{amsthm}
\usepackage{subcaption}
\usepackage{enumitem}
\usepackage{algorithm}
\usepackage{algpseudocode}
\usepackage{algpseudocodex}

\newtheorem{theorem}{Theorem}
\newtheorem{definition}{Definition}
\newtheorem{example}{Example}
\newtheorem{remark}{Remark}
\usepackage{float}
\usepackage{newfloat}
\usepackage{listings}
\usepackage{amsfonts}
\DeclareCaptionStyle{ruled}{labelfont=normalfont,labelsep=colon,strut=off} 
\floatstyle{ruled}
\newfloat{listing}{tb}{lst}{}
\floatname{listing}{Listing}

\title{Unlocking Dynamic Inter-Client Spatial Dependencies: A Federated Spatio-Temporal Graph Learning Method for Traffic Flow Forecasting}
\author{
    Feng Wang\textsuperscript{\rm 1,2},
    Tianxiang Chen\textsuperscript{\rm 1,2},
    Shuyue Wei\textsuperscript{\rm 1,3},
    Qian Chu\textsuperscript{\rm 1,2},\\
    Yi Zhang\textsuperscript{\rm 4}\footnote{Corresponding authors.},
    Yifan Sun\textsuperscript{\rm 5}\footnotemark[\value{footnote}],
    Zhiming Zheng\textsuperscript{\rm 1,2}
}
\affiliations{
    \textsuperscript{\rm 1}Beijing Advanced Innovation Center for Future Blockchain and Privacy Computing\\
    \textsuperscript{\rm 2}School of Artificial Intelligence, Beihang University, China\\
    \textsuperscript{\rm 3}School of Computer Science and Engineering, Beihang University, China\\
    \textsuperscript{\rm 4}School of Mathematics, Renmin University of China, China\\
    \textsuperscript{\rm 5}Center for Applied Statistics and School of Statistics, Renmin University of China, China\\
    \{fengwang, txchen, weishuyue, qianchu\}@buaa.edu.cn, 
    \{ethanzhang, sunyifan\}@ruc.edu.cn,
    zzheng@pku.edu.cn
    
}

\usepackage{bibentry}

\usepackage{amsmath} 
\usepackage{amsthm}
\usepackage{algorithm}
\usepackage{algpseudocode}
\usepackage{multirow}
\usepackage{ragged2e}
\usepackage{pdfpages}
\usepackage{graphicx} 
\usepackage{booktabs}

\begin{document}

\maketitle

\begin{abstract}
Spatio-temporal graphs are powerful tools for modeling complex dependencies in traffic time series. However, the distributed nature of real-world traffic data across multiple stakeholders poses significant challenges in modeling and reconstructing inter-client spatial dependencies while adhering to data locality constraints. Existing methods primarily address static dependencies, overlooking their dynamic nature and resulting in suboptimal performance. In response, we propose \underline{\textbf{Fed}}erated \underline{\textbf{S}}patio-\underline{\textbf{T}}emporal \underline{\textbf{G}}raph with  \underline{\textbf{D}}ynamic 
Inter-Client Dependencies (FedSTGD), a framework designed to model and reconstruct dynamic inter-client spatial dependencies in federated learning. FedSTGD incorporates a federated nonlinear computation decomposition module to approximate complex graph operations. This is complemented by a graph node embedding augmentation module, which alleviates performance degradation arising from the decomposition. These modules are coordinated through a client-server collective learning protocol, which decomposes dynamic inter-client spatial dependency learning tasks into lightweight, parallelizable subtasks. Extensive experiments on four real-world datasets demonstrate that FedSTGD achieves superior performance over state-of-the-art baselines in terms of RMSE, MAE, and MAPE, approaching that of centralized baselines. Ablation studies confirm the contribution of each module in addressing dynamic inter-client spatial dependencies, while sensitivity analysis highlights the robustness of FedSTGD to variations in hyperparameters.
\end{abstract}


\section{Introduction}
\begin{figure*}[t] 
    \centering 
        \includegraphics[width=\textwidth]{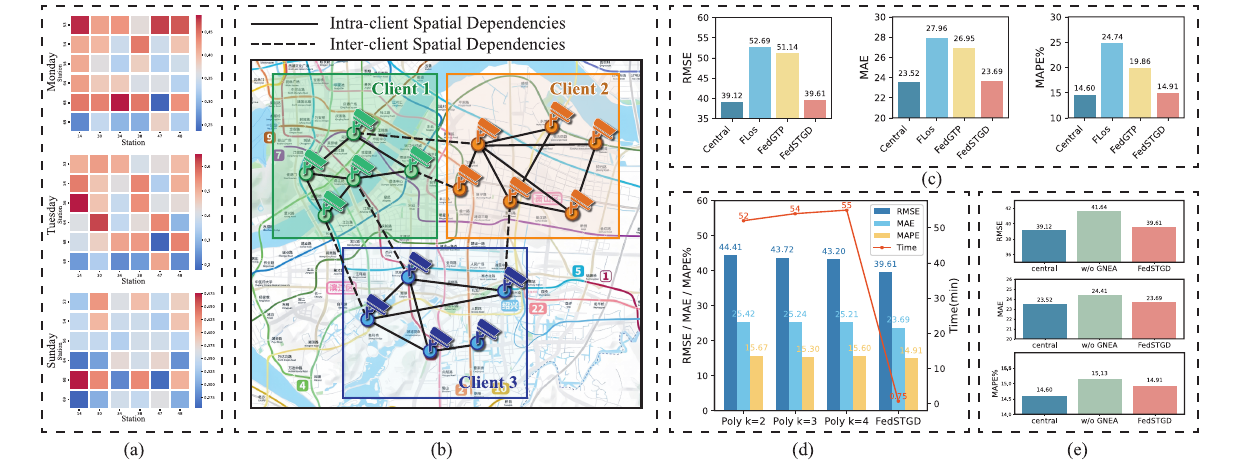} 
    \caption{Experimental results on the HZMetro dataset. (a) Visualization of dynamic inter-client spatial dependencies; (b) Schematic illustration of intra-client and inter-client spatial dependencies; (c) Performance comparison between FedSTGD and other mainstream methods; (d) Comparison of Federated Nonlinear Computation Decomposition in FedSTGD against AdptPoLU in other mainstream methods; (e) Ablation study on Graph Node Embedding Augmentation module.} 
    \label{intro} 
\end{figure*}

Traffic Flow Forecasting (TFF) is a cornerstone of Intelligent Transportation Systems. The performance of TFF critically depends on the modeling of complex spatial dependencies within transportation systems. These dependencies are naturally encoded as graphs, where nodes denote sensors and edges signify their spatial relationships. Due to their capacity for capturing such spatial dependencies, graph neural networks have become a dominant approach for TFF~\cite{fang2021spatial,li2021spatial, kong2024spatio}.

Spatial dependencies in transportation systems are typically dynamic. To validate this, we examine passenger flow data from the Hangzhou Metro, and observe variations in spatial dependencies between weekdays and weekends, and even within weekdays (Figure~\ref{intro}(a)). Due to the abundant information they contain, dynamic spatial dependencies offer potential to enhance the accuracy of TFF models.

The modeling of dynamic spatial dependencies in real-world applications is often hindered by data fragmentation across diverse stakeholders, such as governmental agencies and private enterprises, collectively referred to as clients \cite{FASTGNN}. This fragmentation is exacerbated by stringent data privacy regulations, such as the General Data Protection Regulation~\cite{GDPR2016}, which restrict data aggregation. Therefore, the distributed nature of traffic data introduces a critical distinction between intra-client and inter-client dynamic spatial dependencies~\cite{ge2024fedaga, FedGTP, zhou2024fault}, as illustrated in Figure~\ref{intro}(b). While dynamic intra-client dependencies can be modeled locally, the data fragmentation renders the modeling of dynamic inter-client spatial dependencies intractable. 

Federated Learning~\cite{mcmahan2017communication} has become a foundational paradigm for collaborative training under data locality, attracting broad research interest across domains~\cite{FCS25_FedLLM_Survey}.
However, existing federated graph learning models either overlook inter-client spatial modeling (e.g., FLoS~\cite{FLoS}) or treat these dependencies as static (e.g., FedGTP~\cite{FedGTP}). Such limitations curtail model expressivity and, in turn, predictive performance. To demonstrate this, we evaluate representative models on the HZMetro dataset (Figure~\ref{intro}(c)). Our proposed FedSTGD, which explicitly models dynamic inter-client spatial dependencies, achieves significant gains, approaching the centralized baselines. However, modeling such dependencies in federated learning confronts \textbf{three key challenges}.

First, efficiently decomposing inter-client nonlinear operations is challenging. Modeling such dependencies requires complex nonlinear operations over distributed client data, which cannot be performed in federated settings due to data locality constraints. Existing methods often rely on polynomial approximations~\cite{chrysos2021deep,fan2023expressivity,FedGTP}, but this entails projections into high-dimensional spaces where dimensionality scales exponentially with polynomial degree, leading to high overhead. Our empirical evaluation on HZMetro dataset (Figure~\ref{intro}(d)) reveals that this method incurs high overhead with minimal performance gains, while ours entails negligible overhead.

Secondly, the representational capacity of graph node embeddings may be insufficient in federated settings. This limitation stems from the decomposition of inter-client nonlinear operations, which compromises the expressive power of the embeddings and results in degraded federated performance. We confirm this limitation through an ablation study on the HZmetro dataset (Figure~\ref{intro}(e)), where a variant without our embedding augmentation falls short of centralized performance, while its inclusion yields performance close to that of centralized baselines.

Third, executing inter-client graph operations remains a formidable challenge. Modeling these dependencies requires computations that span multiple clients, necessitating the decomposition of such operations into parallelizable sub-tasks suitable for distributed execution.

To this end, we introduce FedSTGD ({\textbf{Fed}erated {\textbf{S}}patio- {\textbf{T}}emporal {\textbf{G}}raph with  {\textbf{D}}ynamic 
Inter-Client Dependencies), a novel framework designed to enhance TFF by explicitly modeling dynamic inter-client spatial dependencies. FedSTGD consists of three key components: (1) {\it Module of federated non-linear computation decomposition} decomposes non-linear computations across clients; (2) {\it Module of graph node embedding augmentation}  enriches the representational capacity of node embeddings; (3) {\it Module of client-server collective learning} coordinates dynamic inter-client spatial dependencies modeling between clients and central server. The contributions of our work are summarized as follows:
\begin{itemize} 

\item  We make the first attempt to model and reconstruct dynamic inter-client spatial dependencies in a federated spatio-temporal graph model by introducing a novel recovery mechanism under the data locality constraint.

\item We propose a novel framework, dubbed FedSTGD. This framework employs a federated nonlinear computation decomposition module to approximate complex inter-client nonlinear computations as lightweight, client-side operations. To address resulting approximation errors, we integrate a graph node embedding augmentation module for enhancing local representations. On this basis, a client-server collective learning module enables the recovery of dynamic inter-client spatial dependencies by decomposing the tasks into lightweight, parallelizable subtasks amenable to distributed execution across clients.
\item  Extensive evaluations on four real-world datasets show that  FedSTGD delivers state-of-the-art performance, outperforming diverse baseline methods and closely approximating the performance of centralized baselines. Ablation studies confirm the effectiveness of individual components, while hyperparameter sensitivity analyses affirm the framework's robustness.
\end{itemize}

\section{Related Work}
We review related work on federated spatio-temporal graph-based TFF from three key perspectives: methods that rely on predefined graphs, publicly accessible graphs, and that focus on modeling inter-client spatial dependencies.

\subsubsection{Predefined Graph} 
This line of research leverages predefined graph structures to model inter-client and intra-client dependencies. For instance, FLoS~\cite{FLoS} and MFVSTGNN~\cite{MFVSTGNN} use connectivity matrices to characterize spatial dependencies. However, such approaches yield only rough representations of spatio-temporal dependencies, thereby constraining their predictive performance.

\subsubsection{Publicly Accessible Graphs} This line of research leverages publicly accessible graphs to model inter-client and intra-client dependencies. For instance, CNFGNN~\cite{Cross} extracts cross-node dependencies at the server level through predefined graph topologies, while FedSTN~\cite{FedSTN} separately captures long- and short-term spatio-temporal patterns within a shared road network. Similarly, FedAGCN~\cite{FedAGCN} and FCGCN~\cite{Short} assume a globally shared spatial topology and employ community detection for subgraph partitioning; CTFL~\cite{Efficient} extends this paradigm by assuming identical graph structures across clients. Although these methods partially capture inter-client spatial dependencies, their dependence on global topology assumptions often proves impractical in real-world settings.

\subsubsection{Modeling of Inter-Client Spatial Dependency}  
This line of research focuses on modeling inter-client spatial dependencies in federated learning. Existing methods approximate these dependencies in various ways: FASTGNN~\cite{FASTGNN} introduces noise from differential privacy mechanism; DSTGCRN~\cite{pham2025federated} and FedSTG~\cite{zhang2025fedstg} employ graph-aware neighborhood aggregation; FUELS~\cite{liu2025personalized} applies contrastive dual semantic alignment; and FedGTP~\cite{FedGTP} reconstructs static spatial relations. 
However, these methods either overlook the dynamic nature of inter-client spatial dependencies or achieve only partial approximations.

\section{Preliminaries}
In this section, we define the key components and objectives of our study.

\begin{definition}
A spatio-temporal graph for a traffic network at time $t$ is denoted by $G^t=\left(V, E, A^t\right)$, where $V$ is the set of $N = |V|$ nodes, $E$ is the set of edges reflecting connectivity, and $A^t\in\mathbb{R}^{N\times N}$ is a dynamic adjacency matrix.
\end{definition}

\begin{definition}
The graph signal for $G^t=\left(V, E, A^t\right)$ at time $t$ is denoted by $X^t \in \mathbb{R}^{N \times d}$, where $d$ is the feature dimensionality per node.
\end{definition}

\begin{definition}
Spatio-temporal graph-based TFF is to learn a function $f(\cdot; \boldsymbol{\theta})$ that maps $T_{\text{in}}$ input graph signals $\mathcal{X} = (X^{t-T_{\text{in}}+1}, \dots, X^{t})$ and graph structures $\mathcal{G} = (G^{t-T_{\text{in}}+1}, \dots, G^{t})$ to $T_{\text{out}}$ future graph signals $\mathcal{Y} = (X^{t+1}, \dots, X^{t+T_{\text{out}}})$. The parameters $\boldsymbol{\theta}$ are optimized by minimizing a loss function $\mathcal{L}$ over the training data distribution $\mathcal{D}$:
\begin{equation}
\boldsymbol{\theta}^* = \underset{\boldsymbol{\theta}}{\operatorname{argmin}} \ \mathbb{E}_{(\mathcal{X}, \mathcal{Y}) \sim \mathcal{D}} \left[ \mathcal{L}\left(f\left(\mathcal{G}, \mathcal{X}; \boldsymbol{\theta}\right), \mathcal{Y}\right) \right].
\end{equation}
\end{definition}

An example of a spatial-temporal graph model for TFF is the Time-aware Graph Convolutional Recurrent Network (TGCRN) proposed by~\cite{ma2024learning}.

\begin{example} The TGCRN model operates as follows:
\begin{align} 
\label{eq: Anu}  
&\mathcal{A}_{\nu}=\langle E_{\nu},E_{\nu}^{\top}\rangle,\\  
&\eta_{\tau}^{t}=\langle E_{\tau}^{t}, E_{\tau}^{t-1\top}\rangle,\\  
&\mathcal{A}_\rho= \tanh(\langle X^{(t)},X^{(t)\top}\rangle),\\ \label{eq: At}\displaybreak
& \mathcal{A}^t = (\mathbf{1}_{N\times N}+\alpha\sigma(\mathcal{A}_{\rho}))\odot(\mathcal{A}_{\nu}+\eta_\tau^t\cdot \mathbf{1}_{N\times N}),\\ 
&\hat{E}^t = [E_\nu\| (\mathbf{1}_{N\times 1}\otimes E_{\tau}^t)], \label{eq:embed}\\ 
&z^t = \sigma(\mathcal{A}^t I_1^t  \hat{E}^t  W_z + \hat{E}^t b_z), \label{eq:G}\\
&r^t = \sigma(\mathcal{A}^t I_1^t \hat{E}^t W_r + \hat{E}^t b_r), \label{eq:R}\\
&\hat{h}^t =  \tanh(\mathcal{A}^t I_2^t \hat{E}^t W_{\hat{h}} + \hat{E}^t b_{\hat{h}}), \label{eq:U}\\
& h^{t} = (1 - z^{t}) \odot h^{t-1} + z^{t} \odot \hat{h}^{t}. \label{eq:h}
\end{align}
Here: $\odot$ is the Hadamard product; $\otimes$ is the Kronecker product; $\mathbf{1}_{p\times q}$ denotes the all-ones matrix; $I^t_1 = [X^t\| h^{t-1}]$ and $I^t_2 = [X^t\| r^t \odot h^{t-1}]$ are the gated recurrent unit (GRU) inputs; $E_v \in \mathbb{R}^{N \times d_N}$ is the learnable graph node embeddings; $E_\tau^t \in$ $\mathbb{R}^{1 \times d_T}$ is the learnable time vectors; $\mathcal{A}_\nu, \mathcal{A}_\rho \in \mathbb{R}^{N \times N}$ are the self-learned adjacency matrix and the periodic discriminant matrix; $\eta_\tau^t$ is a scalar trend factor; $\sigma(\cdot)$ is the sigmoid activation; $\alpha$ is a hyperparameter; $(W_z, W_r, W_{\hat{h}})$ and $(b_z, b_r, b_{\hat{h}})$ are learnable weight matrices and bias vectors, respectively.
\end{example}

The spatial-temporal graph-based TFF can be extended to a federated setting with a set of clients $\mathcal{C}=$ $\left\{C_1, \ldots, C_M\right\}$ that partition the global graph, where each client $C_i$ possesses a local graph signal $X_i^t \in \mathbb{R}^{N_i \times d}$ and a subgraph $G_i^t=\left(V_i, E_i, A_{i i}^t\right)$, comprising $N_i=\left|V_i\right|$ nodes, intra-client edges $E_i$, and an adjacency matrix $A_{i i}^t \in \mathbb{R}^{N_i \times N_i}$.

\begin{definition}
Federated spatio-temporal graph-based TFF is to learn a function $f(\cdot; \boldsymbol{\theta})$ that takes, for each client $C_i$, $T_{in}$ graph signals $\mathcal{X}_i = (X_i^{t-T_{in}+1}, \dots, X_i^{t})$ and graph structures $\mathcal{G}_{ii} = (G_{ii}^{t-T_{in}+1}, \dots, G_{ii}^{t})$   as input and predicts $T_{out}$ future graph signals $\mathcal{Y}_i = (X_i^{t+1}, \dots, X_i^{t+T_{out}})$. The parameters $\{\boldsymbol{\theta}_1,\ldots,\boldsymbol{\theta}_M\}$, encompassing both global and private parameters, are optimized by minimizing a loss function $\mathcal{L}$ across the clients, weighted by their node sizes:
\begin{equation}
\begin{aligned}
&\{\boldsymbol{\theta}_1^*, \ldots,\boldsymbol{\theta}_M^*\}=\\
&\underset{\boldsymbol{\theta}_1,\ldots, \boldsymbol{\theta}_M}{\operatorname{argmin}} \sum_{i=1}^M \frac{N_i}{N}\mathbb{E}_{( \mathcal{X}_i, \mathcal{Y}_i) \sim \mathcal{D}_i} \left[ \mathcal{L}\left(f(\mathcal{G}_{ii}, \mathcal{X}_i; \boldsymbol{\theta}), \mathcal{Y}_i\right) \right]
\end{aligned}
\end{equation}
where $\mathcal{D}_i$ is the training data distribution for client $C_i$.
\end{definition}

\begin{figure*}[t] 
    \centering 
        \includegraphics[width=\textwidth]{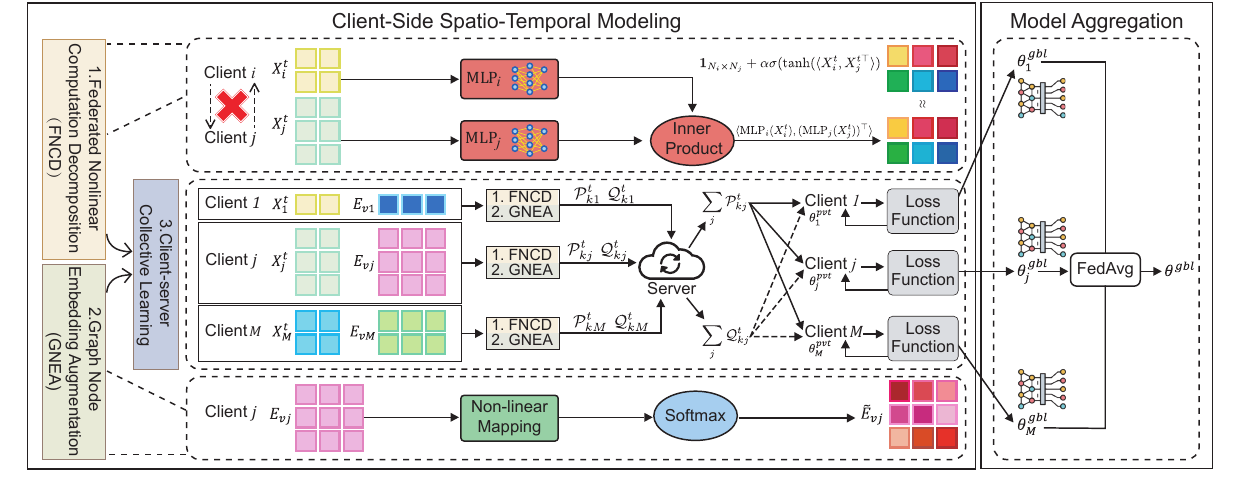} 
    \caption{Overview of the Proposed FedSTGD Framework} 
    \label{frame2} 
\end{figure*}

\section{Methodology}
In this section, we introduce the FedSTGD framework. We start with the federated formation of FedSTGD, then describe three modules for client-side spatio-temporal modeling, followed by the federated model training procedure, and conclude with time and communication complexity analysis. The architecture is shown in Figure \ref{frame2}, with pseudocode provided in Appendix A of the supplementary material.

\subsection{Federated Formulation of FedSTGD.} 
FedSTGD leverages the TGCRN model as its backbone, noted for its state-of-the-art prediction accuracy via learning of dynamic spatial dependencies. In the following, we first describe the client-side partitioning of model components, followed by a concise overview of client-side spatio-temporal modeling and server-side model aggregation.

\subsubsection{1. Client-Side Data, Embedding, and Spatial Dependency Partitioning.} To reformulate the model defined in Eqs.~\eqref{eq: Anu}-\eqref{eq:h} to accommodate federated learning, we partition the model's node features, embeddings, GRU inputs, and adjacency matrix across the clients set $\mathcal{C}$. Specifically, the node features $X^t$, node embeddings $E_\nu$, time embedding $\hat{E}^t$, and the GRU inputs $I_k^t\ (k=1,2)$ are partitioned into client-specific blocks: $\{X_j^t\}_{j=1}^M$, $\{E_{\nu j}\}_{j=1}^M$, $\{\hat{E}_j^t\}_{j=1}^M$, and $\{I_{kj}^t\}_{j=1}^M$ respectively. Additionally, the time-dependent adjacency matrix is partitioned into $\mathcal{A}^{t} = [\mathcal{A}_{ij}^{t}]_{i,j=1}^{M}$, where the diagonal blocks $\mathcal{A}_{ii}^t$ represent \textbf{dynamic intra-client spatial dependencies}, and the off-diagonal blocks $\mathcal{A}_{ij}^t$ ($i \neq j$) encode \textbf{dynamic inter-client spatial dependencies}. 




\subsubsection{2. Client-Side Spatio-Temporal Modeling.}  Building upon the aforementioned partitioning and in accordance with the model specified in Eqs.~\eqref{eq: Anu}--\eqref{eq:h}, we formulate the client-side spatio-temporal modeling as follows:
\begin{align} \label{eq:Li}
    &L_{ki}^t = \sum_{j=1}^{M}\big\{\big[( \mathbf{1}_{N_i\times N_j}+\alpha \sigma(\tanh(\langle X_{i}^{t}, X_{j}^{t  \top}\rangle)  \\\nonumber
&\hspace{1cm}\odot (\langle E_{\nu i},E_{\nu j}^{\top}\rangle+\eta_{\tau i}^t \cdot  \mathbf{1}_{N_i\times N_j})\big] \cdot I_{kj}^t\big\},\ \   \\ \label{eq:zi} 
    &z_i^t = \sigma(L_{i1}^t\hat{E}^t_i W_{zi} + \hat{E}^t_i b_{zi}), \\ 
    &r_i^t = \sigma(L_{i1}^t\hat{E}^t_i W_{ri} + \hat{E}_i^t b_{ri}), \\ 
    &\hat{h}_i^t = \tanh(L_{i2}^t\hat{E}^t_i W_{\hat{h}i} + \hat{E}_i^t b_{\hat{h}i}), \\\label{eq:hi}
    &h_{i}^t = (1 - z_{i}^t) \odot h_{i}^{t-1} + z_{i}^t \odot \hat{h}_{i}^t.
\end{align}
where $W_{zi}$, $W_{ri}$, $W_{\hat{h}i}$, $b_{zi}$, $b_{ri}$, $b_{\hat{h}i}$, and $\eta_{\tau i}^t$ are client-side global learnable parameters; $E_{\nu i}$ is  client-side private learnable parameters. 

\begin{remark} Learning dynamic inter-client spatial dependencies in FedSTGD is \textbf{non-trivial}. This is because learning such dependencies requires inter-client data operations compounded by nonlinear operations (Eq.~\eqref{eq:Li}), which can not be done directly under data locality constraints. Therefore, we will introduce \textbf{three dedicated modules} in the subsequent analysis to address this issue.
\end{remark}

\subsubsection{3. Server-side Model Aggregation.} The server optimizes the global model parameters through the FedAvg algorithm.

\subsection{Federated Nonlinear Computation Decomposition}
In this module, we address the client-level decomposition of nonlinear operations, i.e.,
\begin{align}\label{eq:nonlinear}
\mathbf{1}_{N_i \times N_j}+\alpha\sigma(\tanh(\langle X_i^t,X_j^{t \top}\rangle).
\end{align}
Computing Eq.~\eqref{eq:nonlinear} requires cross-client inner products before applying nonlinearity. This process raises privacy concerns, as it shares data across clients to compute $\langle X_i^t, X_j^t \rangle$. To mitigate this, we try to leverage the one-way property of nonlinear functions, where the forward mapping from data $X_i$ to its transformed representation is efficient, but the inverse mapping is computationally infeasible or noisy, thus protecting against data inference \cite{tian2024towards}. However, the non-commutative nature of these operations renders the direct reversal of the nonlinear transformation and inner product infeasible.

Inspired by the universal approximation theorem~\cite{lu2021learning}, which establishes that the multi-layer perceptron (MLP) can approximate any continuous function with arbitrary precision given sufficient layers and neurons, we propose a federated nonlinear computation decomposition module. In this module, each client locally trains an MLP, and the inner product $\langle \operatorname{MLP}_i(X_i^t), ({\operatorname{MLP}}_j(X_j^t))^{\top} \rangle$ is formulated to approximate $\mathbf{1}_{N_i \times N_j} + \alpha \sigma(\tanh(\langle X_i^t, X_j^{t \top} \rangle))$. This approximation supports the subsequent decomposition of federated computations. Formally, this relationship is expressed as:
\begin{equation}\label{eq:mlp}
\begin{aligned}
&\mathbf{1}_{N_i \times N_j}+\alpha\sigma(\tanh(\langle X_i^t,X_j^{t \top}\rangle) \\
&\hspace{2cm}\approx  \langle\operatorname{MLP}_i(X_i^t), (\operatorname{MLP}_j(X_j^t))^{\top}\rangle.
\end{aligned} 
\end{equation}
Here, $\operatorname{MLP}_i$ represents an $L$-layer MLP with public learnable parameters $W_{\operatorname{MLP}_i}$ and $b_{\operatorname{MLP}_i}$. During training, these parameters are iteratively optimized to enhance the approximation accuracy of $\operatorname{MLP}_i(X_i^t)$ as specified in Eq.~\eqref{eq:mlp}.

\subsection{Graph Node Embedding Augmentation}
In this module, we propose a graph node embedding augmentation strategy to mitigate the degradation in graph representation capacity induced by the previous federated nonlinear computation decomposition. Formally, let $E_{vi} \in \mathbb{R}^{N \times d}$ denote the node embedding for client $i$. Our augmentation module unfolds in two key steps:

\subsubsection{1. Non-Linear Mapping.} We apply a nonlinear transformation to the embedding matrix $E_{v_i}$ to enhance its feature expressiveness:
\begin{equation}
E_{vi}^{\prime}=\mathcal{F}\left(E_{vi}; W_{NL}\right),
\end{equation}
where $\mathcal{F}(\cdot)$ is a nonlinear mapping function (e.g., an MLP, kernel function, or attention mechanism), ${W}_{NL} \in \mathbb{R}^{d \times d^{\prime}}$ are global learnable parameters in $\mathcal{F}(\cdot)$, and $d^{\prime}$ denotes the output dimensionality. This step injects additional representational capacity into the graph node embedding.

\subsubsection{2. Softmax Normalization.} To eliminate scale discrepancies across the augmented graph node embeddings of different clients, we normalize the transformed embedding matrix row-wise using the softmax function:
\begin{equation}\label{eq:gne}
\tilde{E}_{vi}=\operatorname{Softmax}(E_{vi}^{\prime})
\end{equation}
where the softmax operation is performed along the feature dimension, ensuring that each row sums to one.

\subsection{A Client-server Collective Learning Protocol}

In this module, we propose a client-server collective learning protocol to facilitate the distributed execution of client-side spatio-temporal modeling, as formulated in Eqs.~\eqref{eq:Li}--\eqref{eq:hi}. To this end, we first substitute Eqs.~\eqref{eq:mlp} and \eqref{eq:gne} into Eq.~\eqref{eq:Li}, resulting in the following reformulation:
\begin{align}\label{eq:li1}\small
&\tilde{L}_{ki}^t=   \\\nonumber
&  \underbrace{\sum_{j=1}^{M} \left[\langle\operatorname{MLP}_i(X_i^t), (\operatorname{MLP}_j(X_j^t))^{\top}\rangle \odot (\eta_\tau^t\cdot \mathbf{1}_{N_i \times N_j}) \right] \cdot I_{kj}^t}_{P_{ki}^t}     \\ \nonumber
    & + \underbrace{\sum_{j=1}^{M} \left[\langle\operatorname{MLP}_i(X_i^t), (\operatorname{MLP}_j(X_j^t))^{\top}\rangle \odot 
    \langle \tilde{E}_{\nu i}, \tilde{E}_{\nu j}^{\top} \rangle \right] \cdot I_{kj}^t}_{Q_{ki}^t}.
\end{align}
Consequently, the evaluation of Eq.~\eqref{eq:li1} reduces to the computation of $P_{ki}^t$ and $Q_{ki}^t$.

The first component, $P_{ki}^t$, can be reformulated as:
\begin{equation}\label{eq:li3}
P_{ki}^t = \eta_{\tau i}^{t}\cdot \Big[ \operatorname{MLP}_i(X_i^t) \cdot \sum_{j=1}^{M}
((\operatorname{MLP}_j(X_j^t))^{\top} \cdot I_{kj}^t)
\Big].
\end{equation}
Therefore, the computation of $P_{ki}^t$ proceeds as follows:
\begin{itemize}
\item  \textbf{Step} 1. Each client $C_j$ locally computes $\mathcal{P}_{kj}^t = (\operatorname{MLP}_j(X_j^t))^{\top} \cdot I_{kj}^t$ and sends it to the server. 
\end{itemize}
\begin{itemize}
\item \textbf{Step} 2. The server aggregates these terms by computing $\sum_{j=1}^{M} \mathcal{P}_{kj}^t$ and broadcasts the sum back to each client. 
\end{itemize}
\begin{itemize}
\item \textbf{Step} 3. Each client $C_i$ calculates $P_{ki}^t = \eta_{\tau i}^t \cdot \operatorname{MLP}_i(X_i^t) \cdot \sum_{j=1}^{M} \mathcal{P}_{kj}^t$.
\end{itemize}
The component $Q_{ki}^t$ is more challenging to compute in a distributed fashion due to the coupling of inner and Hadamard product operations across different clients' feature matrices and node embeddings. To address this, we introduce an algebraic transformation that decouples the client-side computations. This transformation is formalized in Theorem \ref{THM1}.
\begin{theorem}\label{THM1}
Let $w_{i1}, w_{i2}, \ldots, w_{id}$ be the columns of $\operatorname{MLP}_i(X_i^t) \in \mathbb{R}^{N_i \times d}$ and $v_{i1}, v_{i2}, \ldots, v_{id_N}$ be the columns of $\tilde{E}_{v i} \in \mathbb{R}^{N_i \times d_N}$. Define  $\Gamma: \mathbb{R}^{N_i \times d} \times \mathbb{R}^{N_i \times d_N} \rightarrow \mathbb{R}^{N_i \times(d_N \cdot d)}$ as
\begin{align*}
&\Gamma(\operatorname{MLP}_i(X_i^t), \tilde{E}_{v i})=\big[w_{i1} \odot v_{i1}, \cdots, w_{i1} \odot v_{id_N},\\
&w_{i2} \odot v_{i1}, \cdots, w_{i2} \odot v_{id_N}, \cdots, w_{id} \odot v_{i1}, \cdots, w_{id} \odot v_{id_N}\big].
\end{align*}
Then, we have:
\begin{align*}
&\langle\operatorname{MLP}_i(X_i^t), (\operatorname{MLP}_j(X_j^t))^{\top}\rangle \odot(\tilde{E}_{v i} \cdot \tilde{E}_{v j}^{\top})\\
&\hspace{1.5cm}=\Gamma(\operatorname{MLP}_i(X_i^t), \tilde{E}_{v i}) \cdot \Gamma(\operatorname{MLP}_j(X_j^t), \tilde{E}_{v j})^{\top} .
\end{align*}
\end{theorem}
\noindent The proof of Theorem \ref{THM1} is shown in Appendix B in supplementary materials.

Based on Theorem \ref{THM1}, we reformulate component $Q_{ki}^{t}$ as:
\begin{equation}\label{eq:li2}
\begin{aligned}
&Q_{ki}^{t} =\Gamma(\operatorname{MLP}_i(X_i^t), \tilde{E}_{v i})\\
&\hspace{2cm}\times\sum_{j=1}^{M} \Big( \Gamma(\operatorname{MLP}_j(X_j^t), \tilde{E}_{v j})^{\top}\cdot I_{kj}^t \Big).
\end{aligned}
\end{equation}
This permits the computation of $Q_{ki}^t$ as follows: 
\begin{itemize}
\item \textbf{Step} 1. Each client $C_j$  computes $\mathcal{Q}_{kj}^t = \Gamma(\operatorname{MLP}_j(X_j^t), \tilde{E}_{\nu j})^{\top}\cdot I_{kj}^t$ and sends it to the server. 
\item \textbf{Step} 2. The server aggregates these terms to form $\sum_{j=1}^M \mathcal{Q}_{kj}^t$ and broadcasts the sum.
\item \textbf{Step} 3. Each client $C_i$ calculates $Q_{ki}^{t} =\Gamma(\operatorname{MLP}_i(X_i^t), \tilde{E}_{v i})\cdot \sum_{j=1}^M \mathcal{Q}_{kj}^t.$
\end{itemize}
Following the steps above, client $i$ computes $\tilde{L}_{ki}^t$ while adhering to data locality constraints. Subsequently, client $i$ evaluates Eqs. \eqref{eq:zi}-\eqref{eq:hi}, thereby finishing the client-side spatio-temporal modeling.

\subsection{Training Process}
After the client-side spatio-temporal modeling, each client evaluates the loss function and optimizes both global and private model parameters using stochastic gradient descent. The private parameters remain local, whereas the global parameters are uploaded to the server for aggregation via the FedAvg algorithm.

\subsection{Time and Communication Complexity Analysis}
\subsubsection{Time Complexity.} The time complexity of the FedSTGD is $O(N \cdot d_N \cdot d^2)$, which scales linearly with the graph size $N$, indicating efficient performance for large-scale graphs.


\subsubsection{Communication Complexity.} The communication complexity of FedSTGD is given by $O(|\theta| \cdot M \cdot R_g + (M \cdot d_N \cdot d^2) \cdot R_l)$, where $R_g$ is the number of global rounds and $R_l$ is the number of local rounds per client. The first term reflects model aggregation across clients, while the second captures client-side spatio-temporal modeling.

\section{Experiments}
In this section, we evaluate the performance of FedSTGD on a series of experiments over four real-world datasets, which are summarized to answer the following research questions:
\begin{itemize}
\item \textbf{RQ1}: How does the TFF performance of FedSTGD compare to various baseline models?
\item \textbf{RQ2}: What is the contribution of each designed module to the overall model performance?
\item \textbf{RQ3}: How do variations in hyperparameters influence the performance of the model?
\end{itemize}

\subsection{Experimental Setup}
\subsubsection{Datasets}
To evaluate the performance of FedSTGD, we utilize four real-world datasets: HZMetro \cite{Physical}, SHMetro \cite{Physical}, NYC-Bike\footnote {https://www.citibikenyc.com/system-data}, and NYC-Taxi\footnote{https://www.nyc.gov/site/tlc/about/tlc-trip-record-data.page}. Each dataset offers unique traffic flow dynamics. Detailed descriptions of these datasets are provided in Table \ref{tab: data}.
\begin{table}[h]
\setlength{\tabcolsep}{3pt}
\begin{tabular}{@{}lllll@{}}
\toprule
Datasets & Nodes & Size    & TimeStep & TimeRange           \\ \midrule
HZMetro  & 80    & 2.35M   & 15 min        & Jan 1-25, 2019       \\
SHMetro  & 288   & 811.8M   & 15 min        & Jul 1-Sep 30, 2016   \\
NYC-Bike & 250   & 30.7M     & 30 min        & Apr 1-Jun 30, 2016   \\
NYC-Taxi & 266   & 35M     & 30 min        & Apr 1-Jun 30, 2016   \\ \bottomrule
\end{tabular}
\caption{Descriptions of the Datasets.}
\label{tab: data}
\end{table}


\subsubsection{Implementation Details}
We conducted all experiments on a computational platform equipped with an Intel Xeon Gold 5215 CPU (2.50 GHz) and an NVIDIA A100 40 GB GPU. The Adam optimizer was employed, configured with an initial learning rate of $10^{-3}$, an $L_2$ regularization penalty of $10^{-4}$, and a learning rate decay factor of 0.3 applied at epoch milestones [5, 20, 40, 70, 90]. A batch size of 16 was used throughout. For the graph-based model, the hyperparameter $\alpha$ was set to 0.3, with node embedding dimension $d_N$ and temporal embedding dimension $d_T$ configured to 64. The federated nonlinear computation decomposition module employed an MLP architecture, structured with an initial layer mapping inputs to four neurons, followed by a ReLU activation function, and a final Softmax layer to produce output probabilities. The same MLP configuration was adopted for the nonlinear function in the graph node embedding augmentation.

\subsubsection{Evaluation Metrics} Performance was evaluated using the following metrics: Root Mean Square Error (RMSE), Mean Absolute Error (MAE), Mean Absolute Percentage Error (MAPE). 


\subsubsection{Baselines.}
We compare FedSTGD with the following 9 baselines belonging to three classes. (1) \textbf{predefined graph-based methods} (FLoS~\cite{FLoS}, MFVSTGNN~\cite{MFVSTGNN}); (2) \textbf{publicly accessible graph-based methods} (CTFL~\cite{Efficient}, FCGCN~\cite{Short}); (3) \textbf{inter-client spatial dependency recovery methods} (FASTGNN~\cite{FASTGNN}, FedGTP~\cite{FedGTP}, FedSTG~\cite{zhang2025fedstg}, FUELS~\cite{liu2025personalized}, DSTGCRN~\cite{pham2025federated}). Details are provided in Appendix C in the supplementary materials.

\subsection{Performance Comparison (RQ1)}
\begin{table*}[htbp]
  \centering
  \setlength{\tabcolsep}{4pt}
    \begin{tabular}{cccc|ccc|ccc|ccc}
    \toprule
    \multirow{2}[4]{*}{Method} & \multicolumn{3}{c}{HZMetro} & \multicolumn{3}{c}{SHMetro} & \multicolumn{3}{c}{NYC Bike} & \multicolumn{3}{c}{NYC Taxi} \\
\cmidrule{2-13}          
& RMSE  & MAE   & MAPE\%  
& RMSE  & MAE   & MAPE\%  
& RMSE  & MAE   & MAPE\% 
& RMSE  & MAE   & MAPE\% \\
    \midrule
    MFSTGNN  
    &58.90&34.03&22.50
    &78.96&35.85&19.35
    &2.95&1.86&19.90       
    &19.08&9.49&18.92\\
    FLoS  
    &52.69 &27.96 &20.93       
    &75.38 &35.31 &19.97       
    &2.89&1.81&18.92
    &17.79&10.23&18.83\\\midrule
    FCGCN 
    &54.67 &29.06 &23.73       
    &79.33 &36.61 &22.04       
    &2.87&1.62&18.70  
    &18.77&10.63&18.56\\
    CTFL 
    &54.41&28.99 &21.60 &84.30&41.79&20.30
    &2.89&1.65&18.71    
    &17.03 &8.74&16.94\\ \midrule
    FASTGNN &68.37&36.45&23.42
    &101.22&49.74&24.96
    &2.90&1.68&18.68    
    &19.17&10.26&19.87     \\
    FedSTG 
    &56.31&29.76&22.01
    &68.26&31.91&22.41
    &2.86&1.81&18.81
    &15.49&7.50&18.63\\
    DSTGCRN 
    &56.12&29.13&22.76
    &71.01&33.60&22.98
    &3.04&1.89&19.87
    &26.15&9.32&20.65
    \\
    FUELS
    &54.98&30.81&22.49
    &66.57&29.23&21.67
    &2.86&1.72&18.92
    &12.19&6.56&17.90\\
    FedGTP &51.14&26.95&19.86 
    &64.92&26.88&18.04      
    &2.85&1.72&18.63
    &10.74&5.23&17.52\\\midrule
    \textbf{FedSTGD}
    &\textbf{39.61}&\textbf{23.69}& \textbf{14.91} &\textbf{50.08}&\textbf{24.24}&\textbf{17.36}       &\textbf{2.73}&\textbf{1.61}&\textbf{18.52} 
    &\textbf{9.03}&\textbf{4.93} &\textbf{17.17}\\
    \bottomrule
    \end{tabular}%
\caption{Comparison of performance on the traffic flow forecasting task between FedSTGD and baselines.}
  \label{tab:CompareBaseline}%
\end{table*}%

Table \ref{tab:CompareBaseline} presents the results from federated graph-based baselines. The best results are highlighted in bold. Several key conclusions can be drawn:
\begin{itemize}
\item  FedSTGD consistently outperforms all competing methods across every evaluation metric and dataset examined.
\item Among the predefined graph-based methods, several approaches rely on predefined connectivity matrices to characterize spatial dependencies, resulting in poor forecasting performance.
\item Among the publicly accessible graph-based methods, many overlook privacy concerns by treating graph structure information as publicly available. However, the inherent limitations of the graph models employed lead to consistently poor performance.
\item Among the inter-client spatial dependency modeling methods, some of them attempt to approximate the inter-client spatial dependency, yet empirical results indicate insufficient approximation capabilities. The FedGTP framework fully reconstructs inter-client spatial dependencies, but its focus on static dependencies constrains its predictive accuracy, rendering it inadequate for capturing dynamic spatial interactions.
\end{itemize}


\subsection{Ablation Study (RQ2)}
To assess the impact of individual components within FedSTGD, we conduct an ablation study by constructing several variants of the framework:
\begin{itemize}
\item \textbf{w/o GNEA}: This variant ablates the modeling of graph node embedding augmentation. 
\item \textbf{w/o -,All}: This variant ablates all the modeling of inter-client spatial dependencies, both static and dynamic.
\item \textbf{w/o Dyn,Dyn}: This variant ablates the modeling of inter-client and intra-client dynamic spatial dependencies.
\item \textbf{w/o Dyn,All}: This variant ablates the modeling of intra-client dynamic spatial dependencies and inter-client spatial dependencies.
\item \textbf{w/o All,All}: This variant ablates all the modeling of inter-client spatial dependencies.
\end{itemize}

\begin{table}[htbp]\small
\setlength{\tabcolsep}{3pt}
\begin{tabular}{@{}lllllll@{}}
\toprule
\multirow{2}{*}{Model} & \multicolumn{3}{c}{HZMetro} & \multicolumn{3}{c}{NYC Taxi} \\ \cmidrule(l){2-7} 
                       & RMSE    & MAE    & MAPE\%   & RMSE    & MAE    & MAPE\%    \\ \midrule
w/o GNEA 
& 41.64 & 24.41 & 15.13 &9.98&5.36&19.03  \\\midrule
w/o All,All
&62.57&27.93&21.94
&17.91&10.43&20.82\\
w/o -,All
&43.23 &25.21 &15.60 
&9.58&5.24&18.21\\
w/o Dyn,Dyn
&51.14 &26.95 &19.86 
&10.74&5.23&14.12\\
w/o Dyn,All
&52.66 &26.50 &21.67 
&9.87&5.49 &18.91\\ \midrule
FedSTGD&39.61 &23.69&14.91 
&9.03&4.93&17.17\\
\bottomrule
\end{tabular}
\caption{Ablation study on HZMetro and NYC Taxi datasets.}
\label{AS}
\end{table}

\begin{figure}[htbp] 
    \centering 
    \includegraphics[width=0.47\textwidth]{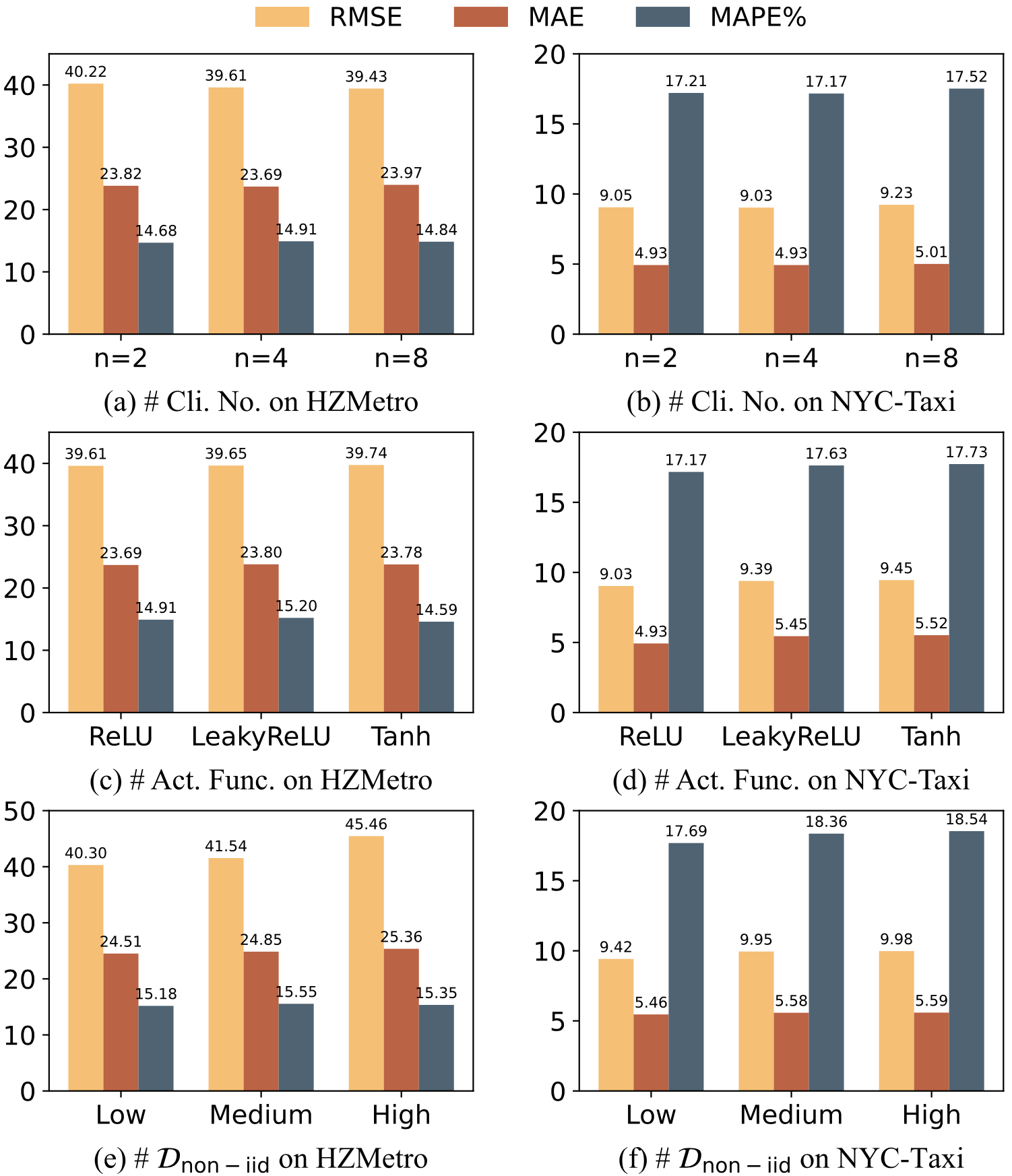} 
    \caption{Parameters Sensitivity Analysis on HZMetro and NYC-Taxi Datasets.}
    \label{SA}
\end{figure}

Table \ref{AS} presents the comparison of FedSTGD and its variants on HZMetro and NYC-Taxi datasets. From this comparison, we can draw several conclusions: (1) FedSTGD consistently achieves the best performance relative to its variants, underscoring the effectiveness of its full configuration. (2) The variant ``w/o GNEA" underperforms FedSTGD, demonstrating the critical role of the graph node embedding augmentation module. (3) The variant ``w/o All,All" underperforms FedSTGD, demonstrating the critical role of spatial modeling.  (4) The variant ``w/o -,All" underperforms FedSTGD, emphasizing the importance of modeling inter-client spatial dependencies.
(5) The variants ``w/o Dyn,Dyn" and ``w/o Dyn,All" underperform FedSTGD, underscoring the necessity of modeling dynamic spatial dependencies at both inter-client and intra-client levels.

\subsection{Sensitivity Analysis (RQ3)}
Figure \ref{SA} presents the results of a hyperparameter sensitivity analysis for our FedSTGD framework on the HZMetro and NYC-Taxi datasets. This analysis involves varying the number of clients across the set $\{2, 4, 8\}$, evaluating different activation functions within the MLP—including ReLU, LeakyReLU, Tanh—and considering three distinct levels of data heterogeneity (i.e., $\mathcal{D}_{non-iid}$). Several key insights can be drawn from these experiments: (1) The model's performance demonstrates overall robustness as the number of clients and the degree of data heterogeneity increase, though with a modest decline. (2) Variations in the MLP activation functions have a negligible impact on the model's outcomes. These findings emphasize the stability of FedSTGD under diverse hyperparameter settings, underscoring its suitability for scalable applications in federated learning environments.

\section{Conclusion}
In this paper, we propose FedSTGD, a novel federated spatio-temporal graph learning framework that effectively captures dynamic inter-client spatial dependencies while adhering to data locality constraints. Comprehensive experiments demonstrate the superior performance of FedSTGD over state-of-the-art baselines, underscoring the critical role of reconstructing dynamic spatial dependencies in federated learning. Ablation studies validate the efficacy of each module in modeling dynamic inter-client spatial dynamics, whereas sensitivity analyses affirm the robustness of FedSTGD to hyperparameter variations.

\section{Acknowledgments}
This work is supported by the fundamental research funds for the central universities, and the research funds of Renmin University of China (23XNL014).

\bibliography{aaai2026}

%

\appendix
\onecolumn
\section{Supplementary Material}
\subsection*{Appendix A: Pseudocode for FedSTGD.}

\begin{algorithm}[htbp]
\caption{FedSTGD Framework}
\label{alg:fedstgd_simplified}
\begin{algorithmic}[1]
\Require  \textbf{Public initial:} model parameters $\Theta^{(0)}=\{W_z^{(0)},W_r^{(0)},W_{\hat h}^{(0)},W_{MLP}^{(0)},W_{NL}^{(0)},b_z^{(0)},b_r^{(0)},b_{\hat h}^{(0)},b_{MLP}^{(0)}\}$ and temporal embedding $E_{\tau}^{(0)}$.

\Statex \textbf{Private initial:} node embedding $\{E_{v_i}^{(0)}\}_{i=1}^M$.
  \Statex Number of global rounds $R_g$, number of local rounds $R_l$.

\Ensure For each client $C_i$, the learned parameters $\{\Theta_i,\,E_{vi},\,E_{\tau i}\}$.

\State Server initializes global parameters: $\Theta_g \gets \Theta^{(0)},E_{\tau g} \gets E_{\tau}^{(0)}$.

\For{each client $C_i$ in parallel}
  \State Initialize private node embeddings: $E_{v_i} \gets E_{v_i}^{(0)}$.
\EndFor

\For{global round $r_g = 1,\dots,R_g$}
  \State Server sends $\Theta_g$ and $E_{\tau}$ to each client $C_i$, setting $\Theta_i \gets \Theta_g$, $E_{\tau_i}\gets E_{\tau g}$.
  \For{local round $r_l = 1,\dots,R_l$}
    \State Perform client-side spatial–temporal modeling (Eqs. (22)–(23)).
    \State Update local parameters via gradient descent: $\{\Theta_i,E_{vi},E_{\tau i}\}$.
  \EndFor
  \State Each $C_i$ uploads its updated $\Theta_i$ and $E_{\tau_i}$ to the server.
  \State Server aggregates $\{\Theta_i,E_{\tau i}\}$ with \textsc{FedAvg} to update $\Theta_g$ and $E_{\tau g}$.
\EndFor

\State \Return For each client $C_i$: $\{\Theta_i,\,E_{vi},\,E_{\tau i}\}$.
\end{algorithmic}
\end{algorithm}

\subsection*{Appendix B: Proof of Theorem 1.}
\begin{theorem}
Let $w_{i1}, w_{i2}, \ldots, w_{id}$ be the columns of $\operatorname{MLP}_i(X_i^t) \in \mathbb{R}^{N_i \times d}$ and $v_{i1}, v_{i2}, \ldots, v_{id_N}$ be the columns of $E_{v i} \in \mathbb{R}^{N_i \times d_N}$. Define  $\Gamma: \mathbb{R}^{N_i \times d} \times \mathbb{R}^{N_i \times d_N} \rightarrow \mathbb{R}^{N_i \times(d_N \cdot d)}$ as
\begin{align*}
&\Gamma(\operatorname{MLP}_i(X_i^t), \tilde{E}_{v i})=\big[w_{i1} \odot v_{i1}, w_{i1} \odot v_{i2}, \cdots, w_{i1} \odot v_{id_N}, w_{i2} \odot v_{i1}, \cdots, w_{id} \odot v_{id_N}\big]
\end{align*}
Then, we have:
\begin{align*}
&\langle\operatorname{MLP}_i(X_i^t), (\operatorname{MLP}_j(X_j^t))^{\top}\rangle \odot(\tilde{E}_{v i} \cdot \tilde{E}_{v j}^{\top})=\Gamma(\operatorname{MLP}_i(X_i^t), \tilde{E}_{v i}) \cdot \Gamma(\operatorname{MLP}_j(X_j^t), \tilde{E}_{v j})^{\top} .
\end{align*}
\end{theorem}

\noindent\textbf{Proof of Theorem 1.} To establish the theorem, we first introduce the necessary notation. Let $A_i = \operatorname{MLP}_i(X_i) \in \mathbb{R}^{N_i \times d}$, with column vectors denoted as $w_{i1}, w_{i2}, \dots, w_{id}$. Let $B_i = \tilde{E}_{vi} \in \mathbb{R}^{N_i \times d_N}$, with column vectors $v_{i1}, v_{i2}, \dots, v_{id_N}$. Similarly, let $A_j = \operatorname{MLP}_j(X_j) \in \mathbb{R}^{N_j \times d}$, with column vectors $w_{j1}, w_{j2}, \dots, w_{jd}$, and $B_j = \tilde{E}_{vj} \in \mathbb{R}^{N_j \times d_N}$, with column vectors $v_{j1}, v_{j2}, \dots, v_{jd_N}$. For an arbitrary matrix $A$, $[A]_{s,t}$ denotes the element in the $s$-th row and $t$-th column of $A$. For an arbitrary column vector $\alpha$, $(\alpha)_t$ denotes the $t$-th element of $\alpha$.

Consider the left-hand side, $(A_i\cdot A_j^\top) \odot (B_i\cdot B_j^\top)$. The $(s, t)$-th element is given by
\begin{align*}
[(A_i\cdot A_j^\top) \odot (B_i\cdot B_j^\top)]_{s,t} = [A_i\cdot A_j^\top]_{s,t} \cdot [B_i\cdot B_j^\top]_{s,t}.
\end{align*}
We compute each term as follows:
\begin{align*}
[A_i\cdot A_j^\top]_{s,t} = \sum_{k=1}^d [A_i]_{s,k}\cdot [A_j]_{t,k} = \sum_{k=1}^d (w_{ik})_s (w_{jk})_t,
\end{align*}
Similarly,
\begin{align*}
[B_i\cdot B_j^\top]_{s,t} = \sum_{l=1}^{d_N} [B_i]_{s,l}\cdot [B_j]_{t,l} = \sum_{l=1}^{d_N} (v_{il})_s (v_{jl})_t,
\end{align*}
Therefore,
\begin{align*}
[(A_i\cdot A_j^\top) \odot (B_i\cdot B_j^\top)]_{s,t} &= \left( \sum_{k=1}^d (w_{ik})_s (w_{jk})_t \right) \left( \sum_{l=1}^{d_N} (v_{il})_s (v_{jl})_t \right) \notag \\
&= \sum_{k=1}^d \sum_{l=1}^{d_N} (w_{ik})_s (w_{jk})_t (v_{il})_s (v_{jl})_t.
\end{align*}

Now consider the right-hand side, $\Gamma(A_i, B_i) \Gamma(A_j, B_j)^\top$. The $(s, t)$-th element is
\begin{align*}
[\Gamma(A_i, B_i)\cdot \Gamma(A_j, B_j)^\top]_{s,t} = \sum_{p=1}^{d \cdot d_N} [\Gamma(A_i, B_i)]_{s,p} \cdot [\Gamma(A_j, B_j)]_{t,p},
\end{align*}
where the columns of $\Gamma(A_i, B_i)$ correspond to the index pairs $(k, l)$ for $k=1,\dots,d$ and $l=1,\dots,d_N$ in lexicographical order, with the $p$-th column associated with a fixed $(k, l)$. Thus,
\begin{align*}
[\Gamma(A_i, B_i)]_{s,p} = [w_{ik} \odot v_{il}]_s = (w_{ik})_s (v_{il})_s,
\end{align*}
and
\begin{align*}
[\Gamma(A_j, B_j)]_{t,p} = [w_{jk} \odot v_{jl}]_t = (w_{jk})_t (v_{jl})_t.
\end{align*}
Consequently,
\begin{align*}
[\Gamma(A_i, B_i)\cdot \Gamma(A_j, B_j)^\top]_{s,t} = \sum_{k=1}^d \sum_{l=1}^{d_N} (w_{ik})_s (v_{il})_s (w_{jk})_t (v_{jl})_t,
\end{align*}
establishing the equality between the left- and right-hand sides.

\subsection*{Appendix C: Detailed Descriptions of Baselines.}

\begin{itemize} 
\item MFVSTGNN (Liu et al. 2023)  models inter-client spatial dependencies with multilevel federated learning using semantic and spatial-temporal graphs for intelligent traffic flow forecasting.
\item FLoS (Wang et al. 2022) processes inter-client spatial dependencies via dynamic spatio-temporal GCNs in federated learning for traffic flow prediction.
\item FCGCN (Xia, Jin, and Chen 2023) integrates community detection-based federated learning with a graph convolutional network to construct a short-term traffic flow prediction model.
\item CTFL (Zhang et al. 2022) optimizes inter-client spatial dependencies via client clustering and hierarchical federated learning for communication-efficient graph-based traffic forecasting.
\item FASTGNN  (Zhang et al. 2021) seeks to recover inter-client spatial dependencies by incorporating noise from differential privacy mechanisms.
\item FedSTG (Zhang et al. 2025) addresses inter-client spatial dependencies by applying neighborhood aggregation in federated graph learning to break spatio-temporal data silos in traffic forecasting.
\item DSTGCRN (Pham et al. 2025) manages inter-client spatial dependencies through federated LSTM replacements for GRUs and client-side validation, enhancing dynamic spatiotemporal modeling.
\item FUELS (Liu et al. 2025) captures inter-client spatial dependencies via dual semantic alignment-based contrastive learning in personalized federated settings to address spatio-temporal heterogeneity.
\item FedGTP (Yang et al. 2024) exploits inter-client spatial dependencies through adaptive reconstruction in federated graph-based frameworks for enhanced traffic prediction.
\end{itemize}

\end{document}